\pdfoutput=1

\documentclass[11pt]{article}

\usepackage[]{acl}
\usepackage{bbm}
\usepackage{times}
\usepackage{calc}
\usepackage{graphicx}
\usepackage{tabularx}
\usepackage{multirow}
\usepackage{booktabs}
\usepackage{latexsym}
\usepackage{amsmath}
\usepackage{algorithm}
\usepackage{algpseudocode}
\usepackage{bbm}
\newcommand{\V}[1]{\mathbf{#1}}
\usepackage[T1]{fontenc}
\usepackage{array}
\usepackage[utf8]{inputenc}

\usepackage{microtype}

\newcommand\model{\textsc{LingMess}}

\definecolor{darkspringgreen}{rgb}{0.09, 0.45, 0.27}

\title{\model{}: Linguistically Informed Multi Expert Scorers \\for Coreference Resolution}

\author{Shon Otmazgin\textsuperscript{1} \quad 
        Arie Cattan\textsuperscript{1} \quad
        Yoav Goldberg\textsuperscript{1,2} \quad \\ 
        \textsuperscript{1}Computer Science Department, Bar Ilan University \\ 
        \textsuperscript{2}Allen Institute for Artificial Intelligence \\ 
  {\normalsize\tt  \{shon711,arie.cattan,yoav.goldberg\}@gmail.com} \\
 }

\begin{document}
\maketitle

\begin{abstract}


Current state-of-the-art coreference systems are based on a single pairwise scoring component, which assigns to each pair of mention spans a score reflecting their tendency to corefer to each other. We observe that different kinds of mention pairs require different information sources to assess their score. We present \model{}, a linguistically motivated categorization of mention-pairs into 6 types of coreference decisions and learn a dedicated trainable scoring function for each category. This significantly improves the accuracy of the pairwise scorer as well as of the overall coreference performance on the English Ontonotes coreference corpus and 5 additional datasets.\footnote{The codebase to train and run \model{} is available in \url{https://github.com/shon-otmazgin/lingmess-coref}. Also, our recent \textsc{F-coref} Python package~\citep{otmazgin-etal-2022-f} includes a simple and efficient implementation of \model{} in \url{https://github.com/shon-otmazgin/fastcoref}.}

\end{abstract}
\section{Introduction}

Coreference resolution is the task of clustering textual mentions that refer to the same discourse entity. 
This fundamental task requires many decisions. In this work, we argue that different \emph{kinds} of decisions involve different challenges. To illustrate that, consider the following text:

\emph{``\textbf{Lionel Messi} has won a record seven Ballon d'Or awards. \textbf{He} signed for Paris Saint-Germain in August 2021. ``\textbf{I} would like to thank \textbf{my} family'', said \textbf{the Argentinian footballer}. \textbf{Messi} holds the records for most goals in La Liga''} 

To correctly identify that the pronoun ``He'' refers to ``Lionel Messi'', models need to model the discourse, while linking ``my'' to ``I'' may rely more heavily on morphological agreement. Likewise, linking ``the Argentinian footballer'' to ``Lionel Messi'' requires world knowledge, while linking ``Messi'' to ``Lionel Messi'' may be achieved by simple lexical heuristics. 

Indeed, pre-neural coreference resolution works often considered the types of a mention-pair, either by incorporating this information as model features, or by tailoring specific rules or specific models for each mention pair (see related work section).

However, neural-network based coreference models are all based on a \textit{single} pairwise scorer that is shared for all mention pairs, regardless of the different challenges that needs to be addressed by each pair type \citep{lee-etal-2017-end, lee-etal-2018-higher, joshi-etal-2019-bert, kantor-globerson-2019-coreference, joshi-etal-2020-spanbert, xu-choi-2020-revealing, xia-etal-2020-incremental, toshniwal-etal-2020-learning, thirukovalluru-etal-2021-scaling, kirstain-etal-2021-coreference,  dobrovolskii-2021-word}.

\begin{figure}[t]
    \centering
    \includegraphics[scale=0.359]{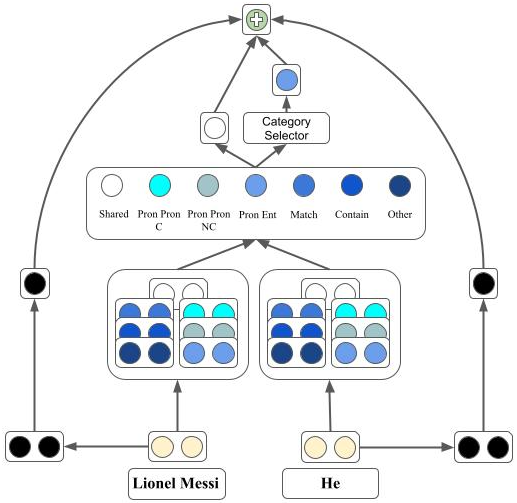}
    \caption{Architecture of our multi expert model. Given two spans \emph{``Lionel Messi''} and \emph{``He''}, we sum four scores: individual mention scores (black), $f_m(\emph{``Lionel Messi''})$, $f_m(\emph{``He''})$, and pairwise scores, shared antecedent score (white) $f_{a}(\emph{``Lionel Messi''}, \emph{``He''})$ and the relevant ``expert'' score (blue) $f_{a}^{\textsc{Pron-Ent}}(\emph{``Lionel Messi''}, \emph{``He''})$.}
    \label{fig:head}
\end{figure}

\begin{table*}[ht]
\small
\centering
\resizebox{\linewidth}{!}{
\begin{tabular}[t]{p{0.14\linewidth}>{\raggedright}p{0.4\linewidth}>{\raggedright\arraybackslash}p{0.39\linewidth}}
\toprule
\textbf{Category} & \textbf{Co-referring example} & \textbf{Non Co-referring example} \\
\midrule

\textsc{Pron-Pron-C} & \emph{A couple of \textbf{my} law clerks were going to ... and \textbf{I} was afraid \textbf{I} was going to...} &  \emph{The Lord God said to \textbf{my} Lord: ``Sit by me at \textbf{my} right side , and I will put your enemies ...} '' \\

\midrule

\textsc{Pron-Pron-NC} & \emph{``\textbf{I} made a similar line and \textbf{I} produced it cheaper'' , \textbf{he} says. } & \emph{\textbf{She} is \textbf{my} Goddess ...} \\

\midrule

\textsc{Ent-Pron} & \emph{\textbf{Spain, Argentina, Thailand and Indonesia} were doing too little to prevent ... across \textbf{their} borders.} & \emph{Tonight, to kick off \textbf{the effort}, CNN will premiere \textbf{its} first prime - time newscast in years.} \\ 

\midrule

\textsc{Match} & \emph{... says Paul Amos, \textbf{CNN} executive vice president for programming. Accordingly, \textbf{CNN} is ... } & \emph{Hertz and Avis can not benefit \textbf{Budget's} programs, '' said Bob Wilson, \textbf{Budget's} vice president ...} \\ 

\midrule

\textsc{Contains} & \emph{He reportedly showed \textbf{DeLay} a videotape that made him weep. \textbf{Tom DeLay} then ...} & \emph{Give \textbf{SEC} authority to halt securities trading, (also opposed by new \textbf{SEC chairman}) ...} \\

\midrule

\textsc{Other} & \emph{They also saw \textbf{the two men who were standing with him}. When \textbf{Moses and Elijah} were leaving ...} & \emph{\textbf{The company} is already working on its own programming ... \textbf{the newspaper} said. } \\

\bottomrule
\end{tabular}}
\caption{Example of each category, taken from Ontonotes development set. We define the categories of mention pairs as follows. \textsc{Pron-Pron-C}: compatible pronouns based on their attributes such as gender, number and animacy (see Appendix~\ref{app:cats} for more details), \textsc{Pron-Pron-NC}: incompatible pronouns, \textsc{Ent-Pron}: a pronoun and another span, 
\textsc{Match}: non-pronoun spans with the same content words, 
\textsc{Contains}: one contains the content words of the other,
\textsc{Other}: all other pairs. Content words exclude stop words, see Appendix~\ref{app:cats} for the full list of stop words.}
\label{tab:categories}
\end{table*}

In this work, we suggest that modeling different mention pairs by different
sub-models (in our case, different learned scoring functions) depending 
on their types is beneficial also for neural models. We identify a set of decisions: (a) linking compatible pronouns (\textsc{Pron-Pron-C}); (b) linking incompatible pronouns (\textsc{Pron-Pron-NC}); (c) linking pronouns to entities (\textsc{Ent-Pron}); (d) linking entities which share the exact lexical form (\textsc{Match}); (e) linking entities where the lexical form of one contains the lexical form of the other (\textsc{Contains}); (f) other cases (\textsc{Other}). Each of these classes is easy to identify deterministically, each contains both positive and negative instances, and each could benefit from a somewhat different decision process. Table~\ref{tab:categories} demonstrates the classes.\footnote{More fine-grained distinctions are of course also possible, but we leave exploration of them to future work.}

We present \textbf{Ling}uistically Informed \textbf{M}ulti \textbf{E}xpert \textbf{S}corer\textbf{s} (\model{}), a coreference model which categorizes each pairwise decision into one of these classes, and learns, in addition to a shared scoring function, also a separate scoring function for each pair type. At inference time, 
for each pair of mentions being scored, we deterministically identify the pair's type, and use the corresponding scoring function.\footnote{For computational efficiency on the GPU, we find it beneficial to compute all the scoring functions and mask away the not needed values.} 

Specifically, we extend the recent \emph{s2e}'s model~\citep{kirstain-etal-2021-coreference} by adding per-category scoring, but the method is general and may work with other coreference models as well. 
As illustrated in Figure~\ref{fig:head}, the final coreference score between two spans is composed---in addition to the individual mention scores---of two pairwise scores: a shared antecedent-compatibility score and an ``expert'' antecedent compatibility score which depends on the linguistic-type of the pair. 

We show that this significantly improves the coreference performance on Ontonotes~\citep{pradhan-etal-2012-conll} and 5 additional datasets. We also inspect the performance of the model for each category separately, showing that some classes improve more than others. This analysis provides a finer-grained understanding of the models and points out directions for future research.

\section{Background: the s2e Model}
\label{sec:bg}

The \emph{s2e} model~\citep{kirstain-etal-2021-coreference} achieves the current best coreference scores among all practical neural models.\footnote{We define ``practical models'' as those that require a constant number transformer-based document encodings per passage, as opposed to a constant number of document encodings per mention. The CorefQA model~\citep{wu-etal-2020-corefqa} achieves a substantially higher score, but requires to run a separate BERT inference for each mention, making it highly impractical. }

Given a sequence of tokens $x_1, \ldots, x_n$, each mention pair $(c, q)$ is scored using a scoring function $F_{\textsc{s}}(c, q)$\footnote{\textsc{s} state for \textsc{shared}} described below, where $c$ is a ``candidate span'' and $q$ is a ``query span'' which appears before $c$ in the sentence. The span encodings are based on contextualized word embeddings obtained by a Longformer encoder, see \citet{kirstain-etal-2021-coreference} for details. These pairwise scores are then used to form coreference chains (see ``inference'' below). 

The scoring function $F_\textsc{s}$ is further decomposed: 
\begin{align*}
F_{\textsc{s}}(c, q) = 
\begin{cases}
f_m(c) + f_m(q) + f_a(c, q) & c \neq \varepsilon \\
0 & c = \varepsilon 
\end{cases}
\end{align*}
where $\varepsilon$ is the null antecedent, and $f_m$ and $f_a$ are parameterized functions, scoring each individual span ($f_m$) and the pairwise interaction ($f_a$).

For each possible mention $q$, the learning objective optimizes the sum of probabilities over the true antecedent $\hat{c}$ of $q$:
\begin{align*}
L_{\textsc{s}}(q) = \log \sum_{\hat{c} \in \mathcal{C}(q) \cap \textsc{gold}(q)}P_{\textsc{s}}(\hat{c} \mid q) 
\end{align*}
where $\mathcal{C}(q)$ is the set of all candidate antecedents\footnote{All spans before $q$ that passed some pruning threshold.} with a null antecedent $\varepsilon$. $\textsc{gold}(q)$ is the set of the true antecedents of $q$.
$P_\textsc{s}(c \mid q)$ is computed as a softmax over $F_\textsc{s}(c, q)$ scores for $c$ values in $\mathcal{C}(q)$:
\begin{align*}
P_{\textsc{s}}(\hat{c} \mid q) = \frac{\exp{F_\textsc{s}(\hat{c}, q)}}{\sum\limits_{c' \in \mathcal{C}(q)} \exp{F_\textsc{s}(c', q)}} 
\end{align*}

\section{\model{}}
\label{sec:lingmess}

Clustering coreferring entities typically involves many different phenomena, which we argue should be addressed in a different manner.
Therefore, our core contribution is proposing to allocate a dedicated scorer $f^t_a(c, q)$ for each phenomenon type $t$, in addition to the shared pairwise scorer $f_a(c, q)$. 
The overall architecture of our model is shown in Figure~\ref{fig:head}.

Concretely, we extend the \emph{s2e} model with six additional antecedent scorers $f_a^{t}$ where $t\in \{\textsc{Pron-Pron-C}, \textsc{Pron-Pron-NC},\textsc{Ent-Pron},$ $\textsc{Match}, \textsc{Contains},\textsc{Other}\}$, the six categories we list in Table \ref{tab:categories}. 

The pairwise scoring function now becomes:
\begin{align*}
F(c, q) =& 
\begin{cases}
f_m(c) + f_m(q) + f(c, q) & c \neq \varepsilon \\ 
0 & c = \varepsilon \\
\end{cases}\\
f(c, q) =& f_a(c, q) + f^{T(c,q)}_{a}(c, q)
\end{align*}
where $T(c, q)$ is a deterministic, rule-based function to determine the category $t$ of the pair $(c, q)$. 
The pairwise scoring function $f(c,q)$ scoring $c$ as the antecedent of $q$, is now composed of a shared scorer $f_a(c, q)$ and an ``expert'' scorer $f^t_a(c, q)$ which differs based on the type of the pair $c, q$.
Each of the seven pairwise scoring functions ($f_a$ and the six $f_a^t$) is parameterized separately using its own set of matrices. The transformer-based encoder and the mention scorer $f_m$ are shared between all the antecedent scorers. See Appendix~\ref{app:lingmess} for full model architecture.

\paragraph{Training} 
For each span $q$, our model optimizes the objective function $L_{\textsc{Coref}}$ over the sum of probabilities of all true antecedents of $q$:
\begin{align*}
L_{\textsc{Coref}}(q) = \log \sum_{\hat{c} \in \mathcal{C}(q) \cap \textsc{gold}(q)}P(\hat{c} \mid q)
\end{align*}
Here, $P(\hat{c} \mid q)$ is a softmax over $F(\hat{c}, q)$ scores, that is our new score function described in Figure~\ref{fig:head}.

\begin{align*}
P(\hat{c} \mid q) = \frac{\exp{F(\hat{c}, q)}}{\sum\limits_{c' \in \mathcal{C}(q)} \exp{F(c', q)}}
\end{align*}

This scorer is also the one used in inference. However, this objective does not explicitly push each category (``expert'') to specialize. For example, for the \textsc{Pron-Pron-C} cases, it would be useful to explicitly train the model to distinguish between the possible antecedents of that type only (without regarding other antecedents), as well as to explicitly distinguish between a pronoun antecedent and a null antecedent. To this end, we extend the training objective by also training each ``expert'' separately:

\begin{align*}
L_{t}(q) = \log \sum_{\hat{c} \in \mathcal{C}_t(q) \cap \textsc{gold}(q)}P_t(\hat{c} \mid q)
\end{align*}

\begin{table*}[!t]
\small
\centering
\resizebox{\textwidth}{!}{
    \begin{tabular}{@{}lcccccccccccccccccc@{}}
    \toprule
    \multicolumn{1}{c}{} & & \multicolumn{3}{c}{MUC} & & \multicolumn{3}{c}{B\textsuperscript{3}} & & \multicolumn{3}{c}{CEAF\textsubscript{$\phi_4$}} & & \multicolumn{3}{c}{LEA}\\
     \cmidrule{3-5} \cmidrule{7-9} \cmidrule{11-13} \cmidrule{15-17}
      & & R & P & F1 & & R & P & F1 & & R & P & F1 & & R & P & F1 & & Avg. F1 \\
    \midrule
    \midrule
    \emph{s2e}  && \textbf{85.2} & 86.6 & 85.9 && 77.9 & 80.3 & 79.1 && 75.4 & 76.8 & 76.1 && 75.8 & 78.3 & 77.0 && 80.3 \\
    \model{}                                && 85.1 & \textbf{88.1} & \textbf{86.6} && \textbf{78.3} & \textbf{82.7} & \textbf{80.5} && \textbf{76.1} & \textbf{78.5} & \textbf{77.3} && \textbf{76.3} & \textbf{80.9} & \textbf{78.5} && \textbf{81.4} \\
    

    \bottomrule
    \end{tabular}
}
\caption{Performance on the test set of the English OntoNotes 5.0 dataset. The averaged F1 of MUC, B\textsuperscript{3}, CEAF\textsubscript{$\phi$} is the main evaluation metric. Our model outperforms the \emph{s2e} model~\citep{kirstain-etal-2021-coreference} by 1.1 CoNLL F1. The performance gain is statistically significant according to a non-parametric permutation test ($p < 0.05$). }
\label{table:results}
\end{table*}
\begin{table}[t!]
\small
\centering
\resizebox{0.49\textwidth}{!}{
\begin{tabular}{@{}lcccc@{}}
\toprule
   & \emph{s2e} & \model{} \\
\midrule
\midrule
WikiCoref~\citep{ghaddar-langlais-2016-wikicoref}       & 59.7 & \textbf{62.6}  \\
GAP~\citep{webster-etal-2018-mind}             & 88.3 & \textbf{89.6}  \\
WinoGender~\citep{rudinger-etal-2018-gender}      & 70.5 & \textbf{77.3}  \\
WinoBias~\citep{zhao-etal-2018-gender}       & 84.3 & \textbf{85.1}  \\
BUG~\citep{levy-etal-2021-collecting-large}             & 72.2 & \textbf{74.6}  \\
\bottomrule
\end{tabular}}
\caption{Performance on the test set of various coreference datasets. The reported metrics are CoNLL F1 for WikiCoref, F1 for GAP and Accuracy for WinoGender, WinoBias and BUG.}
\label{table:other}
\end{table}
\begin{table}[t]
\small
\centering
\resizebox{0.48\textwidth}{!}{
\begin{tabular}{@{}lccccccccccccccccccc@{}}
\toprule
\multicolumn{1}{c}{} &    \multicolumn{3}{c}{\emph{s2e}} & & \multicolumn{3}{c}{\model{}} \\
 \cmidrule{2-4} \cmidrule{6-8} 
   &    P & R & F1 & & P & R & F1  \\
\midrule
\midrule

\textsc{Pron-Pron-C}    &  88.8 & 71.3 & 79.1 & &  88.0 & 85.1 & \bf 86.5  \\ 
\textsc{Pron-Pron-NC}   &  84.2 & 55.8 & 67.1 & &  88.3 & 68.7 & \bf 77.3  \\ 
\textsc{Ent-Pron}       &  78.8 & 68.7 & 73.4 & &  80.4 & 69.8 & \bf 74.7  \\
\textsc{Match}          &  85.6 & 90.2 & 87.8 & &  85.3 & 93.7 & \bf 89.3  \\
\textsc{Contains}       &  72.4 & 80.9 & 76.4 & &  77.4 & 78.9 & \bf 78.1  \\
\textsc{Other}          &  60.1 & 70.2 & 64.7 & &  71.7 & 64.2 & \bf 67.7  \\

\bottomrule
\end{tabular}}

\caption{Pairwise performance by category, on the test set of the English OntoNotes 5.0 dataset. \model{} surpasses the \emph{s2e} model~\citep{kirstain-etal-2021-coreference} for most categories by a substantial margin.}
\label{table:pairwise}
\end{table}

\begin{align*}
P_t(\hat{c} \mid q) = \frac{\exp{F_t(\hat{c}, q)}}{\sum\limits_{c' \in \mathcal{C}_t(q)} \exp{F_t(c', q)}}
\end{align*}

\begin{align*}
F_t(c, q) = 
\begin{cases}
f_m(c) + f_m(q) + f^t_a(c, q) & c \neq \varepsilon \\ 
0 & c = \varepsilon
\end{cases}
\end{align*}

Note that for $L_t(q)$ we replace $\mathcal{C}(q)$ with $\mathcal{C}_t(q)$, considering only the potential antecedents that are compatible with the span $q$ for the given type. For example, for $L_{\textsc{Match}}$ and a span $q$, we will only consider candidates $c$ which appear before $q$ with the exact same content words as $q$.
Our final objective for each mention span $q$ is thus:
\begin{align*}
L(q) &= L_{\textsc{Coref}}(q) + L_{\textsc{Experts}}(q) \\
L_{\textsc{Experts}}(q) &=  \sum_{t} L_{t}(q) + L_{\textsc{s}}(q)
\end{align*}

\paragraph{Inference} At inference time, we compute the score of each mention based on the shared scorer as well as the per-type scorer matching the mention type. We then form the coreference chains by linking each mention $q$ to its most likely antecedent $c$ according to $F(c, q)$. We do not use higher-order inference as it has been shown to have a marginal impact~\citep{xu-choi-2020-revealing}.

\section{Experiments}
\label{sec:experiments}

Our baseline is the \textit{s2e} model trained on the English OntoNotes 5.0 dataset by its authors~\citep{kirstain-etal-2021-coreference}. We train \model{} also on OntoNotes, and evaluate both models on OntoNotes, WikiCoref, GAP, WinoGender, WinoBias and BUG. 
Implementation details are described in Appendix~\ref{app:implementation}.

\paragraph{Performance}
Table~\ref{table:results} presents the performance of \model{} on the test set of Ontonotes. \model{} achieves 81.4 F1 on Ontonotes, while the \emph{s2e} baseline achieves 80.3. Our performance gain is statistically significant according to a non-parametric permutation test ($p$ < 0.05). Additionally, Table~\ref{table:other} shows that \model{} outperforms the \emph{s2e} model on WikiCoref (+2.9) GAP (+1.3), WinoGender (+6.8), WinoBias (+0.8) and BUG (+2.4), indicating a better out-of-domain generalization.

\paragraph{Importance of per-category scoring.}

To assess that the improvement of \model{} is due to the decomposition into our set of categories and not to the added parameters, we do two experiments. First, we train a random baseline, which randomly assigns a category for each pair\footnote{For each pair of mentions $(c, q)$, we take the modulo of the sum of the ASCII code of the last character of the last token of $c$ and $q$.} and obtain similar results as the baseline. Second, we train our model by optimizing only the overall loss $L_{\textsc{Coref}}$ and not $L_{\textsc{Experts}}$. This achieves lower results than the baseline, due to low mention recall. 

In addition to the standard coreference evaluation, we report pairwise performance for each category. Given a mention-pair $(c, q)$, if $F(c, q)$ is higher than 0, we treat it as a positive prediction, otherwise negative.
We then measure precision, recall and F1 based on gold clusters labels. 
Table~\ref{table:pairwise} shows the pairwise performance of the \emph{s2e} model and \model{}. \model{} outperforms \emph{s2e} by a significant margin for all categories (e.g +7.4 F1 for \textsc{Pron-Pron-C}, +10.2 F1 for \textsc{Pron-Pron-NC}, etc.).\footnote{These gains in this pairwise metric are higher than the CoNLL metrics reported in Table~\ref{table:results}, because the CoNLL metrics are based on the final clusters, after aggregation of individual pairwise decisions.} The performance varies across the different categories, suggesting aspects of the coreference problem where future work can attempt to improve.


\paragraph{The importance of the shared scorer.} 
To investigate the role of the shared scorer, we trained the \model{} model with only the per-type pairwise scorers, excluding the shared pairwise scorer $F_{\textsc{s}}(c, q)$ and its accompanying loss term $L_{\textsc{s}}(q)$. This resulted in a significant decrease in performance (-0.9), specifically in the recall of the mention detection component. However, adding the shared scorer was able to mitigate this degradation by balancing the different ``experts'' pairwise scorers.

\section{Related Work}

Many pre-neural works consider the various linguistic phenomena involved in coreference resolution as a different challenge. The early coreference system by \citet{zhou-su-2004-high} divided the antecedents candidates into distinct coreference categories (e.g., Name Alias, Apposition, Definite Noun, and a few more) and defined tailored rules for each category. Later, \citet{lee-etal-2013-deterministic} proposed the multi-sieves deterministic model, where each sieve adds coreference links between mention pairs from a specific linguistic category (e.g string match, compatible pronoun, etc.). \citet{haghighi-klein-2009-simple} performed an error analysis of their coreference model according to different types of antecedent decisions, such as Proper Noun-Pronoun, Pronoun-Pronoun, etc. Based on this analysis, they focus on fixing the pronoun antecedent choices by adding syntactic features. More recently, \citet{lu-ng-2020-conundrums} analyze empirically the performance of neural coreference resolvers on various fine-grained resolution categories of mentions (e.g gendered pronoun vs. 1st and 2nd pronoun). They find that while models perform well on name and nominal mention pairs with some shared content words, they still struggle with resolving pronouns, particularly relative pronouns.


Early supervised statistical models train a feature-based classifier that incorporates the type of antecedent decision (e.g. pronoun-entity, string match) as features at the mention-pair level~\citep{soon-etal-2001-machine, bengtson-roth-2008-understanding, clark-manning-2015-entity, clark-manning-2016-improving}. Subsequently, \citet{denis-baldridge-2008-specialized} demonstrate that training separate classifiers that specialize in particular types of mentions (e.g third person pronouns, speech pronouns, proper names, definite descriptions, and all other) provides significant performance improvements. \citet{lassalle-denis-2013-improving} took that observation a step further and proposed a more advanced method for model specialization by learning to separate types of mention into optimal classes and their proper feature space. 

In our work, we make progress in coreference systems specialization direction, and show that the incorporation of linguistic information is helpful also in the context of end-to-end neural models.

\section{Conclusion}

We present \model{}, a coreference model that significantly improves accuracy by splitting the scoring function into different categories, and routing each scoring decision to its own category based on a deterministic, linguistically informed heuristic. This indicates that while end-to-end training is very effective, linguistic knowledge and symbolic computation can still be used to improve results.




\section*{Limitations}

In this paper, we consider a set of 6 linguistic categories of mention pairs, as listed in Table~\ref{tab:categories}. These categories might not be optimal for the task, while a different set of finer-grained categories may result to a higher performance gain. 
Another aspect that can be considered as a limitation is the computation of the categories for every possible pairs. Although the model considers only the top-scoring spans, this additional computation layer increases training and inference time over the baseline (see Appendix~\ref{app:runtime} for the exact time). Our linguistic heuristics could be improved by, e.g., running a parser and considering head words. However, we chose not to do so in this work as this will further increase runtime.



\section*{Acknowledgements}
This project has received funding from the European Research Council (ERC) under the European Union’s Horizon 2020 research and innovation programme, grant agreement No. 802774 (iEXTRACT). Arie Cattan is partially supported by the PBC fellowship for outstanding PhD candidates in data science.

\bibliographystyle{acl_natbib}
\bibliography{anthology,custom}

\begin{thebibliography}{37}
\expandafter\ifx\csname natexlab\endcsname\relax\def\natexlab#1{#1}\fi

\bibitem[{Bagga and Baldwin(1998)}]{bagga-baldwin-1998-entity-based}
Amit Bagga and Breck Baldwin. 1998.
\newblock \href {https://doi.org/10.3115/980845.980859} {Entity-based
  cross-document coreferencing using the vector space model}.
\newblock In \emph{36th Annual Meeting of the Association for Computational
  Linguistics and 17th International Conference on Computational Linguistics,
  Volume 1}, pages 79--85, Montreal, Quebec, Canada. Association for
  Computational Linguistics.

\bibitem[{Beltagy et~al.(2020)Beltagy, Peters, and
  Cohan}]{Beltagy2020LongformerTL}
Iz~Beltagy, Matthew~E. Peters, and Arman Cohan. 2020.
\newblock Longformer: The long-document transformer.
\newblock \emph{ArXiv}, abs/2004.05150.

\bibitem[{Bengtson and Roth(2008)}]{bengtson-roth-2008-understanding}
Eric Bengtson and Dan Roth. 2008.
\newblock \href {https://aclanthology.org/D08-1031} {Understanding the value of
  features for coreference resolution}.
\newblock In \emph{Proceedings of the 2008 Conference on Empirical Methods in
  Natural Language Processing}, pages 294--303, Honolulu, Hawaii. Association
  for Computational Linguistics.

\bibitem[{Clark and Manning(2015)}]{clark-manning-2015-entity}
Kevin Clark and Christopher~D. Manning. 2015.
\newblock \href {https://doi.org/10.3115/v1/P15-1136} {Entity-centric
  coreference resolution with model stacking}.
\newblock In \emph{Proceedings of the 53rd Annual Meeting of the Association
  for Computational Linguistics and the 7th International Joint Conference on
  Natural Language Processing (Volume 1: Long Papers)}, pages 1405--1415,
  Beijing, China. Association for Computational Linguistics.

\bibitem[{Clark and Manning(2016)}]{clark-manning-2016-improving}
Kevin Clark and Christopher~D. Manning. 2016.
\newblock \href {https://doi.org/10.18653/v1/P16-1061} {Improving coreference
  resolution by learning entity-level distributed representations}.
\newblock In \emph{Proceedings of the 54th Annual Meeting of the Association
  for Computational Linguistics (Volume 1: Long Papers)}, pages 643--653,
  Berlin, Germany. Association for Computational Linguistics.

\bibitem[{Denis and Baldridge(2008)}]{denis-baldridge-2008-specialized}
Pascal Denis and Jason Baldridge. 2008.
\newblock \href {https://aclanthology.org/D08-1069} {Specialized models and
  ranking for coreference resolution}.
\newblock In \emph{Proceedings of the 2008 Conference on Empirical Methods in
  Natural Language Processing}, pages 660--669, Honolulu, Hawaii. Association
  for Computational Linguistics.

\bibitem[{Dobrovolskii(2021)}]{dobrovolskii-2021-word}
Vladimir Dobrovolskii. 2021.
\newblock \href {https://doi.org/10.18653/v1/2021.emnlp-main.605} {Word-level
  coreference resolution}.
\newblock In \emph{Proceedings of the 2021 Conference on Empirical Methods in
  Natural Language Processing}, pages 7670--7675, Online and Punta Cana,
  Dominican Republic. Association for Computational Linguistics.

\bibitem[{Ghaddar and Langlais(2016)}]{ghaddar-langlais-2016-wikicoref}
Abbas Ghaddar and Phillippe Langlais. 2016.
\newblock \href {https://aclanthology.org/L16-1021} {{W}iki{C}oref: An
  {E}nglish coreference-annotated corpus of {W}ikipedia articles}.
\newblock In \emph{Proceedings of the Tenth International Conference on
  Language Resources and Evaluation ({LREC}'16)}, pages 136--142,
  Portoro{\v{z}}, Slovenia. European Language Resources Association (ELRA).

\bibitem[{Haghighi and Klein(2009)}]{haghighi-klein-2009-simple}
Aria Haghighi and Dan Klein. 2009.
\newblock \href {https://aclanthology.org/D09-1120} {Simple coreference
  resolution with rich syntactic and semantic features}.
\newblock In \emph{Proceedings of the 2009 Conference on Empirical Methods in
  Natural Language Processing}, pages 1152--1161, Singapore. Association for
  Computational Linguistics.

\bibitem[{Joshi et~al.(2020)Joshi, Chen, Liu, Weld, Zettlemoyer, and
  Levy}]{joshi-etal-2020-spanbert}
Mandar Joshi, Danqi Chen, Yinhan Liu, Daniel~S. Weld, Luke Zettlemoyer, and
  Omer Levy. 2020.
\newblock \href {https://doi.org/10.1162/tacl_a_00300} {{S}pan{BERT}: Improving
  pre-training by representing and predicting spans}.
\newblock \emph{Transactions of the Association for Computational Linguistics},
  8:64--77.

\bibitem[{Joshi et~al.(2019)Joshi, Levy, Zettlemoyer, and
  Weld}]{joshi-etal-2019-bert}
Mandar Joshi, Omer Levy, Luke Zettlemoyer, and Daniel Weld. 2019.
\newblock \href {https://doi.org/10.18653/v1/D19-1588} {{BERT} for coreference
  resolution: Baselines and analysis}.
\newblock In \emph{Proceedings of the 2019 Conference on Empirical Methods in
  Natural Language Processing and the 9th International Joint Conference on
  Natural Language Processing (EMNLP-IJCNLP)}, pages 5803--5808, Hong Kong,
  China. Association for Computational Linguistics.

\bibitem[{Kantor and Globerson(2019)}]{kantor-globerson-2019-coreference}
Ben Kantor and Amir Globerson. 2019.
\newblock \href {https://doi.org/10.18653/v1/P19-1066} {Coreference resolution
  with entity equalization}.
\newblock In \emph{Proceedings of the 57th Annual Meeting of the Association
  for Computational Linguistics}, pages 673--677, Florence, Italy. Association
  for Computational Linguistics.

\bibitem[{Kirstain et~al.(2021)Kirstain, Ram, and
  Levy}]{kirstain-etal-2021-coreference}
Yuval Kirstain, Ori Ram, and Omer Levy. 2021.
\newblock \href {https://doi.org/10.18653/v1/2021.acl-short.3} {Coreference
  resolution without span representations}.
\newblock In \emph{Proceedings of the 59th Annual Meeting of the Association
  for Computational Linguistics and the 11th International Joint Conference on
  Natural Language Processing (Volume 2: Short Papers)}, pages 14--19, Online.
  Association for Computational Linguistics.

\bibitem[{Lassalle and Denis(2013)}]{lassalle-denis-2013-improving}
Emmanuel Lassalle and Pascal Denis. 2013.
\newblock \href {https://aclanthology.org/P13-1049} {Improving pairwise
  coreference models through feature space hierarchy learning}.
\newblock In \emph{Proceedings of the 51st Annual Meeting of the Association
  for Computational Linguistics (Volume 1: Long Papers)}, pages 497--506,
  Sofia, Bulgaria. Association for Computational Linguistics.

\bibitem[{Lee et~al.(2013)Lee, Chang, Peirsman, Chambers, Surdeanu, and
  Jurafsky}]{lee-etal-2013-deterministic}
Heeyoung Lee, Angel Chang, Yves Peirsman, Nathanael Chambers, Mihai Surdeanu,
  and Dan Jurafsky. 2013.
\newblock \href {https://doi.org/10.1162/COLI_a_00152} {Deterministic
  coreference resolution based on entity-centric, precision-ranked rules}.
\newblock \emph{Computational Linguistics}, 39(4):885--916.

\bibitem[{Lee et~al.(2017)Lee, He, Lewis, and Zettlemoyer}]{lee-etal-2017-end}
Kenton Lee, Luheng He, Mike Lewis, and Luke Zettlemoyer. 2017.
\newblock \href {https://doi.org/10.18653/v1/D17-1018} {End-to-end neural
  coreference resolution}.
\newblock In \emph{Proceedings of the 2017 Conference on Empirical Methods in
  Natural Language Processing}, pages 188--197, Copenhagen, Denmark.
  Association for Computational Linguistics.

\bibitem[{Lee et~al.(2018)Lee, He, and Zettlemoyer}]{lee-etal-2018-higher}
Kenton Lee, Luheng He, and Luke Zettlemoyer. 2018.
\newblock \href {https://doi.org/10.18653/v1/N18-2108} {Higher-order
  coreference resolution with coarse-to-fine inference}.
\newblock In \emph{Proceedings of the 2018 Conference of the North {A}merican
  Chapter of the Association for Computational Linguistics: Human Language
  Technologies, Volume 2 (Short Papers)}, pages 687--692, New Orleans,
  Louisiana. Association for Computational Linguistics.

\bibitem[{Levy et~al.(2021)Levy, Lazar, and
  Stanovsky}]{levy-etal-2021-collecting-large}
Shahar Levy, Koren Lazar, and Gabriel Stanovsky. 2021.
\newblock \href {https://doi.org/10.18653/v1/2021.findings-emnlp.211}
  {Collecting a large-scale gender bias dataset for coreference resolution and
  machine translation}.
\newblock In \emph{Findings of the Association for Computational Linguistics:
  EMNLP 2021}, pages 2470--2480, Punta Cana, Dominican Republic. Association
  for Computational Linguistics.

\bibitem[{Lu and Ng(2020)}]{lu-ng-2020-conundrums}
Jing Lu and Vincent Ng. 2020.
\newblock \href {https://doi.org/10.18653/v1/2020.emnlp-main.536} {Conundrums
  in entity coreference resolution: Making sense of the state of the art}.
\newblock In \emph{Proceedings of the 2020 Conference on Empirical Methods in
  Natural Language Processing (EMNLP)}, pages 6620--6631, Online. Association
  for Computational Linguistics.

\bibitem[{Luo(2005)}]{luo-2005-coreference}
Xiaoqiang Luo. 2005.
\newblock \href {https://aclanthology.org/H05-1004} {On coreference resolution
  performance metrics}.
\newblock In \emph{Proceedings of Human Language Technology Conference and
  Conference on Empirical Methods in Natural Language Processing}, pages
  25--32, Vancouver, British Columbia, Canada. Association for Computational
  Linguistics.

\bibitem[{Moosavi and Strube(2016)}]{moosavi-strube-2016-coreference}
Nafise~Sadat Moosavi and Michael Strube. 2016.
\newblock \href {https://doi.org/10.18653/v1/P16-1060} {Which coreference
  evaluation metric do you trust? a proposal for a link-based entity aware
  metric}.
\newblock In \emph{Proceedings of the 54th Annual Meeting of the Association
  for Computational Linguistics (Volume 1: Long Papers)}, pages 632--642,
  Berlin, Germany. Association for Computational Linguistics.

\bibitem[{Otmazgin et~al.(2022)Otmazgin, Cattan, and
  Goldberg}]{otmazgin-etal-2022-f}
Shon Otmazgin, Arie Cattan, and Yoav Goldberg. 2022.
\newblock \href {https://aclanthology.org/2022.aacl-demo.6} {{F}-coref: Fast,
  accurate and easy to use coreference resolution}.
\newblock In \emph{Proceedings of the 2nd Conference of the Asia-Pacific
  Chapter of the Association for Computational Linguistics and the 12th
  International Joint Conference on Natural Language Processing: System
  Demonstrations}, pages 48--56, Taipei, Taiwan. Association for Computational
  Linguistics.

\bibitem[{Paszke et~al.(2019)Paszke, Gross, Massa, Lerer, Bradbury, Chanan,
  Killeen, Lin, Gimelshein, Antiga, Desmaison, Kopf, Yang, DeVito, Raison,
  Tejani, Chilamkurthy, Steiner, Fang, Bai, and Chintala}]{NEURIPS2019_9015}
Adam Paszke, Sam Gross, Francisco Massa, Adam Lerer, James Bradbury, Gregory
  Chanan, Trevor Killeen, Zeming Lin, Natalia Gimelshein, Luca Antiga, Alban
  Desmaison, Andreas Kopf, Edward Yang, Zachary DeVito, Martin Raison, Alykhan
  Tejani, Sasank Chilamkurthy, Benoit Steiner, Lu~Fang, Junjie Bai, and Soumith
  Chintala. 2019.
\newblock \href
  {http://papers.neurips.cc/paper/9015-pytorch-an-imperative-style-high-performance-deep-learning-library.pdf}
  {Pytorch: An imperative style, high-performance deep learning library}.
\newblock In H.~Wallach, H.~Larochelle, A.~Beygelzimer, F.~d\textquotesingle
  Alch\'{e}-Buc, E.~Fox, and R.~Garnett, editors, \emph{Advances in Neural
  Information Processing Systems 32}, pages 8024--8035. Curran Associates, Inc.

\bibitem[{Pradhan et~al.(2012)Pradhan, Moschitti, Xue, Uryupina, and
  Zhang}]{pradhan-etal-2012-conll}
Sameer Pradhan, Alessandro Moschitti, Nianwen Xue, Olga Uryupina, and Yuchen
  Zhang. 2012.
\newblock \href {https://aclanthology.org/W12-4501} {{C}o{NLL}-2012 shared
  task: Modeling multilingual unrestricted coreference in {O}nto{N}otes}.
\newblock In \emph{Joint Conference on {EMNLP} and {C}o{NLL} - Shared Task},
  pages 1--40, Jeju Island, Korea. Association for Computational Linguistics.

\bibitem[{Rudinger et~al.(2018)Rudinger, Naradowsky, Leonard, and
  Van~Durme}]{rudinger-etal-2018-gender}
Rachel Rudinger, Jason Naradowsky, Brian Leonard, and Benjamin Van~Durme. 2018.
\newblock \href {https://doi.org/10.18653/v1/N18-2002} {Gender bias in
  coreference resolution}.
\newblock In \emph{Proceedings of the 2018 Conference of the North {A}merican
  Chapter of the Association for Computational Linguistics: Human Language
  Technologies, Volume 2 (Short Papers)}, pages 8--14, New Orleans, Louisiana.
  Association for Computational Linguistics.

\bibitem[{Soon et~al.(2001)Soon, Ng, and Lim}]{soon-etal-2001-machine}
Wee~Meng Soon, Hwee~Tou Ng, and Daniel Chung~Yong Lim. 2001.
\newblock \href {https://doi.org/10.1162/089120101753342653} {A machine
  learning approach to coreference resolution of noun phrases}.
\newblock \emph{Computational Linguistics}, 27(4):521--544.

\bibitem[{Thirukovalluru et~al.(2021)Thirukovalluru, Monath, Shridhar, Zaheer,
  Sachan, and McCallum}]{thirukovalluru-etal-2021-scaling}
Raghuveer Thirukovalluru, Nicholas Monath, Kumar Shridhar, Manzil Zaheer,
  Mrinmaya Sachan, and Andrew McCallum. 2021.
\newblock \href {https://doi.org/10.18653/v1/2021.findings-acl.343} {Scaling
  within document coreference to long texts}.
\newblock In \emph{Findings of the Association for Computational Linguistics:
  ACL-IJCNLP 2021}, pages 3921--3931, Online. Association for Computational
  Linguistics.

\bibitem[{Toshniwal et~al.(2020)Toshniwal, Wiseman, Ettinger, Livescu, and
  Gimpel}]{toshniwal-etal-2020-learning}
Shubham Toshniwal, Sam Wiseman, Allyson Ettinger, Karen Livescu, and Kevin
  Gimpel. 2020.
\newblock \href {https://doi.org/10.18653/v1/2020.emnlp-main.685} {Learning to
  {I}gnore: {L}ong {D}ocument {C}oreference with {B}ounded {M}emory {N}eural
  {N}etworks}.
\newblock In \emph{Proceedings of the 2020 Conference on Empirical Methods in
  Natural Language Processing (EMNLP)}, pages 8519--8526, Online. Association
  for Computational Linguistics.

\bibitem[{Ulmer et~al.(2022)Ulmer, Hardmeier, and Frellsen}]{ulmer2022deep}
Dennis Ulmer, Christian Hardmeier, and Jes Frellsen. 2022.
\newblock deep-significance-easy and meaningful statistical significance
  testing in the age of neural networks.
\newblock \emph{arXiv preprint arXiv:2204.06815}.

\bibitem[{Vilain et~al.(1995)Vilain, Burger, Aberdeen, Connolly, and
  Hirschman}]{vilain-etal-1995-model}
Marc Vilain, John Burger, John Aberdeen, Dennis Connolly, and Lynette
  Hirschman. 1995.
\newblock \href {https://aclanthology.org/M95-1005} {A model-theoretic
  coreference scoring scheme}.
\newblock In \emph{Sixth Message Understanding Conference ({MUC}-6):
  Proceedings of a Conference Held in {C}olumbia, {M}aryland, November 6-8,
  1995}.

\bibitem[{Webster et~al.(2018)Webster, Recasens, Axelrod, and
  Baldridge}]{webster-etal-2018-mind}
Kellie Webster, Marta Recasens, Vera Axelrod, and Jason Baldridge. 2018.
\newblock \href {https://doi.org/10.1162/tacl_a_00240} {Mind the {GAP}: A
  balanced corpus of gendered ambiguous pronouns}.
\newblock \emph{Transactions of the Association for Computational Linguistics},
  6:605--617.

\bibitem[{Wolf et~al.(2020)Wolf, Debut, Sanh, Chaumond, Delangue, Moi, Cistac,
  Rault, Louf, Funtowicz, Davison, Shleifer, von Platen, Ma, Jernite, Plu, Xu,
  Le~Scao, Gugger, Drame, Lhoest, and Rush}]{wolf-etal-2020-transformers}
Thomas Wolf, Lysandre Debut, Victor Sanh, Julien Chaumond, Clement Delangue,
  Anthony Moi, Pierric Cistac, Tim Rault, Remi Louf, Morgan Funtowicz, Joe
  Davison, Sam Shleifer, Patrick von Platen, Clara Ma, Yacine Jernite, Julien
  Plu, Canwen Xu, Teven Le~Scao, Sylvain Gugger, Mariama Drame, Quentin Lhoest,
  and Alexander Rush. 2020.
\newblock \href {https://doi.org/10.18653/v1/2020.emnlp-demos.6} {Transformers:
  State-of-the-art natural language processing}.
\newblock In \emph{Proceedings of the 2020 Conference on Empirical Methods in
  Natural Language Processing: System Demonstrations}, pages 38--45, Online.
  Association for Computational Linguistics.

\bibitem[{Wu et~al.(2020)Wu, Wang, Yuan, Wu, and Li}]{wu-etal-2020-corefqa}
Wei Wu, Fei Wang, Arianna Yuan, Fei Wu, and Jiwei Li. 2020.
\newblock \href {https://doi.org/10.18653/v1/2020.acl-main.622} {{C}oref{QA}:
  Coreference resolution as query-based span prediction}.
\newblock In \emph{Proceedings of the 58th Annual Meeting of the Association
  for Computational Linguistics}, pages 6953--6963, Online. Association for
  Computational Linguistics.

\bibitem[{Xia et~al.(2020)Xia, Sedoc, and
  Van~Durme}]{xia-etal-2020-incremental}
Patrick Xia, Jo{\~a}o Sedoc, and Benjamin Van~Durme. 2020.
\newblock \href {https://doi.org/10.18653/v1/2020.emnlp-main.695} {Incremental
  neural coreference resolution in constant memory}.
\newblock In \emph{Proceedings of the 2020 Conference on Empirical Methods in
  Natural Language Processing (EMNLP)}, pages 8617--8624, Online. Association
  for Computational Linguistics.

\bibitem[{Xu and Choi(2020)}]{xu-choi-2020-revealing}
Liyan Xu and Jinho~D. Choi. 2020.
\newblock \href {https://doi.org/10.18653/v1/2020.emnlp-main.686} {Revealing
  the myth of higher-order inference in coreference resolution}.
\newblock In \emph{Proceedings of the 2020 Conference on Empirical Methods in
  Natural Language Processing (EMNLP)}, pages 8527--8533, Online. Association
  for Computational Linguistics.

\bibitem[{Zhao et~al.(2018)Zhao, Wang, Yatskar, Ordonez, and
  Chang}]{zhao-etal-2018-gender}
Jieyu Zhao, Tianlu Wang, Mark Yatskar, Vicente Ordonez, and Kai-Wei Chang.
  2018.
\newblock \href {https://doi.org/10.18653/v1/N18-2003} {Gender bias in
  coreference resolution: Evaluation and debiasing methods}.
\newblock In \emph{Proceedings of the 2018 Conference of the North {A}merican
  Chapter of the Association for Computational Linguistics: Human Language
  Technologies, Volume 2 (Short Papers)}, pages 15--20, New Orleans, Louisiana.
  Association for Computational Linguistics.

\bibitem[{Zhou and Su(2004)}]{zhou-su-2004-high}
GuoDong Zhou and Jian Su. 2004.
\newblock \href {https://aclanthology.org/C04-1075} {A high-performance
  coreference resolution system using a constraint-based multi-agent strategy}.
\newblock In \emph{{COLING} 2004: Proceedings of the 20th International
  Conference on Computational Linguistics}, pages 522--528, Geneva,
  Switzerland. COLING.

\end{thebibliography}

\appendix

\section{Model Architecture}
Given a sequence of tokens $x_1, ..., x_n$ from an input document, a transformer-based (BERT-like) encoder first forms contextualized representation vectors, $\V{x_1}, ..., \V{x_n}$ for each token in the sequence.

\subsection{The \emph{s2e} Model}
\label{app:s2e}

\paragraph{Mention scorer}
Given a span $q=(\V{x_i}, \V{x_j})$, represented by its start and end tokens, the score for $q$ being a mention is defined as follow:
\begin{align*}
\resizebox{1.0\hsize}{!}{
$\V{m}_{s}(\V{x}) = \text{GeLU} (\mathbf{W}_{\V{m}_{s}} \V{x}) \qquad
\V{m}_{e}(\V{x}) = \text{GeLU} (\mathbf{W}_{\V{m}_{e}} \V{x})
$}
\end{align*}
\begin{align*}
f_m(q) &= \V{m}_{s}(\V{x_i}) \cdot \V{v}_{s} + \V{m}_{s}(\V{x_j}) \cdot \V{v}_{e} \\
       &+ \V{m}_{s}(\V{x_i}) \cdot \V{B}_m \cdot \V{m}_{s}(\V{x_j})
\end{align*}
where $\V{m}_{s}(\V{x})$ and $\V{m}_{e}(\V{x})$ are two non-linear functions to obtain start and end representations for each token $x$, and $f_m(q)$ is a biaffine product over these representations.

\paragraph{Antecedent scorer}
Given two spans, $c=(\V{x_i}, \V{x_j})$ and $q=(\V{x_k}, \V{x_l})$, represented by their start and end tokens, the score for $c$ being an antecedent of $q$ is computed as follow:
\begin{align*}
\resizebox{1.0\hsize}{!}{
$\V{a}_{s}(\V{x}) = \text{GeLU} (\mathbf{W}_{\V{a}_{s}} \V{x}) \qquad
\V{a}_{e}(\V{x}) = \text{GeLU} (\mathbf{W}_{\V{a}_{e}} \V{x})
$}
\end{align*}
\begin{align*}
f_a(c, q) &= \V{a}_{s}(\V{\V{x_i}}) \cdot \V{B}_{ss} \cdot \V{a}_{s}(\V{\V{x_k}}) \\
          &+ \V{a}_{e}(\V{\V{x_j}}) \cdot \V{B}_{es} \cdot \V{a}_{s}(\V{\V{x_k}}) \\
          &+ \V{a}_{s}(\V{\V{x_i}}) \cdot \V{B}_{se} \cdot \V{a}_{e}(\V{\V{x_l}}) \\ 
          &+ \V{a}_{e}(\V{\V{x_j}}) \cdot \V{B}_{ee} \cdot \V{a}_{e}(\V{\V{x_l}})
\end{align*}

Similar to the mention scorer, $\V{a}_{s}(\V{x})$ and $\V{a}_{e}(\V{x})$ are two non-linear functions to obtain start/end representations for each token, and $f_a(c, q)$ is a sum of four bilinear functions over the start and end representations of $c$ and $q$.

\subsection{\model{}}
\label{app:lingmess}



\paragraph{Mention scorer} Our mention scorer is the same as \emph{s2e} mention scorer implementation.

\paragraph{Antecedent scorer} 

As mentioned in the paper~(§\ref{sec:lingmess}), in addition to the shared antecedent scorer $f_a(c, q)$, \model{} includes a dedicated antecedent scorer $f^t_a(c, q)$ for each category $t\in \{\textsc{Pron-Pron-C}, \textsc{Pron-Pron-NC},\textsc{Ent-Pron},$ $\textsc{Match}, \textsc{Contains},\textsc{Other}\}$. The overall score for $c=(\V{x_i}, \V{x_j})$ being an antecedent of $q=(\V{x_k}, \V{x_l})$ becomes the sum of the shared scorer and the relevant category ``expert'' scorer:

\begin{align*}
f(c, q) = f_a(c, q) + f^{T(c,q)}_{a}(c, q)
\end{align*}
where $T(c,q)$ is a deterministic function to determine the category $t$ of the pair $(c, q)$. 

For each category $t$, we define two specific non-linear functions to obtain start and end representations ($\V{a}^{t}_{s}(\V{x})$ and $\V{a}^{t}_{e}(\V{x})$) as well as an ``expert'' antecedent scoring function $f^{t}_a(c, q)$:

\begin{align*}
\resizebox{1.0\hsize}{!}{
$\V{a}^{t}_{s}(\V{x}) = \text{GeLU} (\mathbf{W}^{t}_{\V{a}_{s}} \V{x}) \qquad
\V{a}^{t}_{e}(\V{x}) = \text{GeLU} (\mathbf{W}^{t}_{\V{a}_{e}} \V{x})
$}
\end{align*}
\begin{align*}
f^{t}_a(c, q) &= \V{a}^{t}_{s}(\V{\V{x_i}}) \cdot \V{B}^{t}_{ss} \cdot \V{a}^{t}_{s}(\V{\V{x_k}}) \\
          &+ \V{a}^{t}_{e}(\V{\V{x_j}}) \cdot \V{B}^{t}_{es} \cdot \V{a}^{t}_{s}(\V{\V{x_k}}) \\
          &+ \V{a}^{t}_{s}(\V{\V{x_i}}) \cdot \V{B}^{t}_{se} \cdot \V{a}^{t}_{e}(\V{\V{x_l}}) \\ 
          &+ \V{a}^{t}_{e}(\V{\V{x_j}}) \cdot \V{B}^{t}_{ee} \cdot \V{a}^{t}_{e}(\V{\V{x_l}})
\end{align*} 


Overall, our model introduces 6 learnable matrices for each category ($\mathbf{W}^{t}_{\V{a}_{s}}$, $\mathbf{W}^{t}_{\V{a}_{e}}$, $\V{B}^{t}_{ss}$, $\V{B}^{t}_{es}$, $\V{B}^{t}_{se}$, $\V{B}^{t}_{ee}$). The transformer-based encoder and the mention scorer are shared between all the different pairwise scorers. 


\section{Implementation Details}
\label{app:implementation}

\subsection{Hyperparameteres}
\label{app:hyper}

We extend the \emph{s2e}'s implementation based on PyTorch~\citep{NEURIPS2019_9015} and Transformers~\citep{wolf-etal-2020-transformers}. We used the same hyperparameters (e.g learning rate, warmup, etc.) as the \emph{s2e} model except the hidden size of all matrices $W$ and $B$. As our method introduces a dedicated antecedent scoring function $f^t_a$ function for each category $t$, we reduce the size of these matrices from 3072 to 2048 to fit training into memory in our hardware. Similar to the baseline our head method is on top of Longformer-Large~\cite{Beltagy2020LongformerTL}, resulting in a total of 590M learnable parameters (the \emph{s2e} model contains 494M learnable parameters). We used dynamic batching both for training and inference, specifically 5K tokens in a single batch during training and 10K tokens at inference.  
 
\subsection{Evaluation}

As mentioned in the paper~(§\ref{sec:experiments}), we conduct our experiments on the English portion of the OntoNotes corpus~\citep{pradhan-etal-2012-conll}. This dataset contains 2802 documents for training, 343 for development, and 348 for test.

We evaluate our model according to the standard coreference metric: MUC~\citep{vilain-etal-1995-model}, B\textsuperscript{3}~\citep{bagga-baldwin-1998-entity-based}, CEAF\textsubscript{$\phi_4$}~\citep{luo-2005-coreference}, and LEA~\citep{moosavi-strube-2016-coreference} using the official CoNLL coreference scorer.\footnote{\url{https://github.com/conll/reference-coreference-scorers}.} \model{} achieves 81.6 CoNLL F1 on the development set of Ontonotes. Also, Table~\ref{table:pairwise-dev} presents the pairwise performance on the development set for each category. We compute statistical significance with a non-parametric permutation test using \citet{ulmer2022deep}'s implementation. Table~\ref{table:gap} shows that \model{} consistenly outperforms the \emph{s2e} model on GAP.

\begin{table}[t]
\small
\centering
\resizebox{0.48\textwidth}{!}{
\begin{tabular}{@{}lccccccccccccccccccc@{}}
\toprule
\multicolumn{1}{c}{} &    \multicolumn{3}{c}{\citet{kirstain-etal-2021-coreference}} & & \multicolumn{3}{c}{\model{}} \\
 \cmidrule{2-4} \cmidrule{6-8} 
   &    P & R & F1 & & P & R & F1  \\
\midrule
\midrule

\textsc{Pron-Pron-C}    &  91.7 & 77.5 & 84.0 & &  91.7 & 90.2 & \bf 91.0  \\ 
\textsc{Pron-Pron-NC}   &  88.9 & 66.2 & 75.9 & &  90.2 & 81.3 & \bf 85.5  \\ 
\textsc{Ent-Pron}       &  82.0 & 74.1 & \bf 77.9 & &  81.4 & 74.7 & \bf 77.9  \\
\textsc{Match}          &  88.3 & 87.5 & 87.9 & &  88.4 & 92.0 & \bf 90.2  \\
\textsc{Contains}       &  69.1 & 77.2 & 72.9 & &  76.1 & 73.5 & \bf 74.8  \\
\textsc{Other}          &  56.8 & 67.5 & 61.7 & &  70.8 & 64.4 & \bf 67.5  \\

\bottomrule
\end{tabular}}

\caption{Pairwise performance by category, on the dev set of the English OntoNotes 5.0 dataset.}
\label{table:pairwise-dev}
\end{table}
\begin{table}[t!]
\small
\centering
\begin{tabular}{@{}lcccc@{}}
\toprule
   & \textbf{Masc} & \textbf{Fem} & \textbf{Bias} & \textbf{Overall} \\
\midrule
\midrule
\citet{kirstain-etal-2021-coreference}  & 90.6 & 85.8 & 0.95 & 88.3 \\
\model{}                    & \textbf{91.3} & \textbf{87.8} & \textbf{0.96} & \textbf{89.6}  \\
\bottomrule
\end{tabular}
\caption{Performance on the test set of the GAP coreference dataset. The reported metrics are F1 scores.}
\label{table:gap}
\end{table}

\subsection{Runtime and Memory}
\label{app:runtime}
\begin{table}[t]
\small
\centering
\resizebox{0.4\textwidth}{!}{

\begin{tabular}{@{}lcc@{}}
\toprule

\textbf{} & \textbf{Runtime} & \textbf{Memory} \\
\midrule

\citet{kirstain-etal-2021-coreference} & 28 & 5.4 \\

\model{} & 43 & 5.9 \\

\bottomrule
\end{tabular}}
\caption{Inference time(Seconds) and memory(GiB) on 343 docs of OntoNotes development set. Using Dynamic batching, 10K tokens in a single batch. Hardware, NVIDIA Tesla V100 SXM2}
\label{tab:runtime}
\end{table}

Our model was trained for 129 epochs on a single 32GB NVIDIA Tesla V100 SXM2 GPU. The training took 23 hours. At shown in Table~\ref{tab:runtime}, the run-time at inference time in \model{} is longer than in the \emph{s2e} model because of the category selection for every possible pair of mentions. The memory consumption remains quite similar to the baseline.

\section{Determining pair types}
\label{app:cats}

Our method routes each pair of spans to their corresponding category scorer. This decision is based on the linguistic properties of the spans. Given a mention-pair $(c, q)$, we defined a rule based function $T(c, q)$ that determines the category of that pair. If $c$ and $q$ are both pronouns, if they are compatible according to gender, number and animacy (see Table~\ref{tab:comp} for the full list), the metnion pair will be routed to \textsc{Pron-Pron-C}, otherwise \textsc{Pron-Pron-NC}. If $c$ is pronoun and $q$ is a non-pronoun span (or vise-versa) we route the mention-pair to \textsc{Pron-Ent}. 
We route the remaining pairs to their corresponding categories (\textsc{Match}, \textsc{Contains} or \textsc{Other}) by considering only content words, excluding the following stop words: \emph{\{'s, a, all, an, and, at, for, from, in, into, more, of, on, or, some, the, these, those\}}.
Accordingly, the mentions ``\emph{the U.S. and Japan}'' and ``\emph{Japan and the U.S.}'' are considered \textsc{Match}, ``\emph{This lake of fire}'' and ``\emph{the lake of fire}'' are considered \textsc{Contains} and ``\emph{Bill Clinton}'' and ``\emph{The President}'' are considered \textsc{Other}.


\begin{table}[t]
\small
\centering
\resizebox{0.4\textwidth}{!}{

\begin{tabular}{@{}ll@{}}
\toprule

\textbf{ID} & \textbf{Pronouns} \\
\midrule

1 & \emph{I, me, my, mine, myself} \\
\midrule

2 & \emph{you, your, yours, yourself, yourselves} \\
\midrule

3 & \emph{he, him, his, himself}  \\
\midrule

4 & \emph{she, her, hers, herself}  \\
\midrule

5 & \emph{it, its, itself}  \\
\midrule

6 & \emph{we, us, our, ours, ourselves}  \\
\midrule

7 & \emph{they, them, their, themselves} \\
\midrule

8 & \emph{that, this}  \\

\bottomrule
\end{tabular}}
\caption{List of groups of compatible pronouns, pronouns with the same ID are considered as compatible.}
\label{tab:comp}
\end{table}



\end{document}